\documentclass{article}
\usepackage{ijcai20}

\usepackage{times}
\usepackage{soul}
\usepackage{url}
\usepackage{epsfig}
\usepackage{amssymb}
\usepackage{verbatim}
\usepackage{multirow}
\usepackage[hidelinks]{hyperref}
\usepackage[utf8]{inputenc}
\usepackage[small]{caption}
\usepackage{graphicx}
\usepackage{amsmath}
\usepackage{amsthm}
\usepackage{booktabs}
\usepackage{diagbox}
\usepackage{subfigure}
\usepackage{floatrow}
\usepackage{algorithm}
\usepackage{algorithmic}
\newfloatcommand{capbtabbox}{table}[][\FBwidth]
\usepackage{blindtext}

\newcommand{\fig}[1]{Fig.~\ref{#1}}

\newcommand{\citet}[1]{\citeauthor{#1}~\shortcite{#1}}

\title{DropNAS: Grouped Operation Dropout for Differentiable Architecture Search\footnote{Sponsored by Huawei Innovation Research Program.}}

\author{
Weijun Hong$^1$\footnote{This work is done when Weijun Hong worked as an intern at
Huawei Noah’s Ark Lab.}\and
Guilin Li$^2$\and
Weinan Zhang$^{1}$\footnote{The corresponding author is supported by NSFC 61702327.}\and
Ruiming Tang$^2$\and
\\
Yunhe Wang$^2$\and
Zhenguo Li$^2$\And
Yong Yu$^1$
\\
\affiliations
$^1$Shanghai Jiao Tong University, China\\
$^2$Huawei Noah’s Ark Lab, China\\
\emails
wiljohn@apex.sjtu.edu.cn,
liguilin2@huawei.com,
wnzhang@sjtu.edu.cn,
\{tangruiming,yunhe.wang,li.zhenguo\}@huawei.com,
yyu@apex.sjtu.edu.cn
}

\begin{document}

\maketitle

\begin{abstract}
Neural architecture search (NAS) has shown encouraging results in automating the architecture design. Recently, DARTS relaxes the search process with a   differentiable formulation that leverages weight-sharing and SGD where all candidate operations are trained simultaneously. Our empirical results show that such procedure results in the \textit{co-adaption problem} and \textit{Matthew Effect}: operations with fewer parameters would be trained maturely earlier.  This causes two problems: firstly, the operations with more parameters may never have the chance to express the desired function since those with less have already done the job; secondly, the system will punish those underperforming operations by lowering their architecture parameter,  and they will get   smaller loss gradients, which causes \textit{the Matthew Effect}.   In this paper,  we systematically study these problems and propose a novel grouped operation dropout algorithm named DropNAS to fix the problems with DARTS.    Extensive experiments demonstrate that DropNAS solves the above issues and achieves promising performance. Specifically, DropNAS achieves 2.26\% test error on CIFAR-10, 16.39\% on CIFAR-100 and 23.4\% on ImageNet (with the same training hyperparameters as DARTS for a fair comparison). It is also observed that  DropNAS is robust across variants of the DARTS search space. Code is available at \textit{\url{https://github.com/wiljohnhong/DropNAS}}. 
\end{abstract}

\section{Introduction}

With the rapid growth of deep learning  in recent years \cite{krizhevsky2012imagenet,silver2017mastering,chen2019data}, designing high-performance neural network architecture is attaining increasing attention. However, such architecture design processes involve a great amount of human expertise. More recently, automatic neural architecture search (NAS) has been brought into focus and achieves state-of-the-art results on various tasks including image classification \cite{zoph2016neural,yang2019cars}, object detection \cite{zoph2018learning}, and recommender system \cite{liu2020autofis},  outperforming human-designed architectures.

To reduce the evaluation cost of NAS, one promising search strategy is to leverage weight-sharing: a supernet containing all possible subnets is built, each of the subnets is optimized as the supernet trained \cite{bender2018understanding,liu2018darts,cai2018proxylessnas,pham2018efficient}. The target subnet can be evaluated by inheriting the parameters from the supernet, which strikingly cuts down the search cost. DARTS built the supernet by introducing continuous architecture parameter. Using two-level optimization, the architecture parameter was trained alternatively with the network weights. 

Many following works of DARTS have studied whether the architecture parameters can be properly learned with the current framework, including many questioning the convergence problem of the two-level optimization \cite{li2019stacnas,MiLeNAS,zela2019understanding,guo2019single} and the optimization gap induced by proxy network \cite{cai2018proxylessnas,chen2019progressive,li2019stacnas}. However,  very few have been explored about how well the candidate operations trained in the supernet with parameter sharing can reflect their stand-alone performance. \cite{xie2018snas} raised the child network performance consistency problem, and used Gumble trick to improve the child network performance. However, they did not compare the performance rank of child networks estimated from the DARTS or SNAS framework to the true rank obtained by fully training the stand-alone child subnets. Moreover, SNAS did not manage to surpass the performance of DARTS.

In this work, we explore how well each candidate operation is trained in the supernet, and how we can balance between the supernet's overall training stability and the individual training of  each subnet.  In DARTS, all candidate operations are trained simultaneously during the network weight training step. Our empirical results show that this training procedure leads to two problems:
\begin{itemize}
    \item \textit{The Co-adaption Problem}: Operations with fewer parameters would be trained maturely with fewer epochs and express the desired function earlier than those with more parameters.  In such cases, the operations with more parameters may rarely have the chance  to express the desired function. Therefore, those operations which take  longer  to converge may never have the chance to express what they could do if trained in a stand-alone model. This makes the system prefer  operations that are easier to train.
    \item  \textit{The Matthew Effect}: The system will punish those underperforming operations by lowering their architecture parameters and backward smaller loss gradients, which causes the Matthew Effect. The Matthew Effect makes the case even worse for operations to take a longer time to train, where their architecture parameters assign them  low scores at the very early stage of supernet training. \end{itemize}
%In this paper, we systematically study these problems and propose a novel grouped operation dropout algorithm named DropNAS to fix the problems with DARTS.    Extensive experiments demonstrate
%The early works of NAS which apply reinforcement learning \cite{zoph2016neural} or evolution strategy \cite{real2019regularized} as their searching techniques require training each candidate architecture from scratch to obtain feedback to guide the exploration, which cost vast computational resources, and therefore lead to a paradigm shift to searching efficiently on a weight-sharing one-shot model \cite{brock2017smash, liu2018darts, cai2018proxylessnas, pham2018efficient}.
%Currently one popular search strategy of NAS is to search on a weight-sharing model, where a supernet containing all the underlying subnets is built, each of them is optimized as the supernet trained \cite{brock2017smash,liu2018darts,cai2018proxylessnas,pham2018efficient}. The target architecture can be induced by just requiring to train the supernet once, which strikingly cuts down the search cost.

In this paper,  we systematically study these problems and propose a novel grouped operation dropout algorithm named DropNAS to fix the problems with DARTS. The proposed DropNAS largely improves the  DARTS framework, including the most state-of-the-art modified algorithms such as P-DARTS \cite{chen2019progressive} and PC-DARTS \cite{xu2019pc} on various datasets. It can also be verified that many previous differentiable NAS research works, including DARTS \cite{liu2018darts}, SNAS \cite{xie2018snas} and ProxylessNAS \cite{cai2018proxylessnas}, are essentially special cases of DropNAS, inferior to DropNAS with optimized drop path rate. 
 
In our experiments, we first show the architectures discovered by DropNAS achieve 97.74\% test accuracy on CIFAR-10 and 83.61\% on CIFAR-100, without additional training tricks (such as increasing channels/epochs or using auto-augmentation). For a fair comparison with other recent works, we also train the searched architectures with auto-augmentation: DropNAS can achieve 98.12\% and 85.90\% accuracy on CIFAR-10 and CIFAR-100 respectively. When transferred to ImageNet, our approach reaches only 23.4\% test error. We also conduct experiments on variants of the DARTS search space and demonstrate that the proposed strategy can perform consistently well when a different set of operations are included in the search space.

In summary, our contributions can be listed as follows:
\begin{itemize}
\item We systematically studied \textit{the co-adaption problem} of DARTS and present empirical evidence on how the performance of DARTS is degraded by this problem. 
\item We introduce grouped operation dropout, which is previously neglected in differentiable NAS community, to alleviate \textit{the co-adaption problem}, meanwhile maintaining the training stability of the supernet. 
\item We unify various  differentiable NAS approaches including DARTS, ProxylessNAS and SNAS, showing that all of them are special cases of DropNAS and inferior to it.
\item We conduct sufficient experiments which show  state-of-the-art performance on various benchmark datasets, and the found search scheme is robust across various search spaces and datasets.
\end{itemize}

\section{Methodology}

\subsection{DARTS}

In this work, we follow the DARTS framework \cite{liu2018darts}, whose objective is to search for the best cell that can be stacked to form the target neural networks.  Each cell is defined as a directed acyclic graph (DAG) of $N$ nodes $\{x_0, x_1,...,x_{N-1}\}$, each regarded as a layer of neural network. An edge $E_{(i,j)}$ connects the node $x_i$ and node $x_j$, and consists of a set of candidate operations. We denote the operation space as $\mathcal{O}$, and follow the original DARTS setting where there are seven candidate operations \{Sep33, Sep55, Dil33, Dil55, Maxpool, Avepool, Identity, Zero\} in it.

DARTS  replaces the discrete operation selection with a weighted sum of all candidate operations, which can be formulated as:
\begin{align}
\bar{o}^{(i,j)}(x_i) &= \sum_{o\in\mathcal{O}} p^{(i,j)}_o \cdot o^{(i,j)}(x_i) \nonumber
\\& = \sum_{o\in\mathcal{O}} \frac{\exp(\alpha_o^{(i,j)})}{\sum_{o'\in\mathcal{O}}\exp(\alpha_{o'}^{(i,j)})} \cdot o^{(i,j)}(x_i)\label{con:softmax}.
\end{align}
%\begin{equation}
%\bar{o}^{(i,j)}(x_i) =  \sum_{o\in\mathcal{O}} \frac{\exp(\alpha_o^{(i,j)})}{\sum_{o'\in\mathcal{O}}\exp(\alpha_{o'}^{(i,j)})} \cdot o^{(i,j)}(x_i)\label{con:softmax}
%\end{equation}
This formula explains how a feature mapping $\bar{o}^{(i,j)}(x_i)$ on edge $E_{(i,j)}$ is computed from the previous node $x_i$. Here $\alpha^{(i,j)}_o$ is the architecture parameter, and $p^{(i,j)}_o$ represents the relative contribution of each operation $o^{(i,j)}\in{\mathcal{O}}$.  Within a cell, each immediate node $x_j$ is represented as the sum of all the feature mappings on edges connecting to it: $x_j = \sum_{i<j} \bar{o}^{(i,j)}(x_i)$. In this work, we adopt the one-level optimization  for  stability and data efficiency similar as \cite{li2019stacnas}, where the update rule can be easily obtained by applying stochastic gradient descent on both $w$ and $\alpha$.
%\begin{align}
%	\min_{\alpha,w} \quad &\mathcal{L}_{train}(w, \alpha)
%	\\\text{s.t.}\quad &w^*, \alpha^* = \mathop{\arg\min}_w \mathcal{L}_{train}(w,\alpha)
%\end{align}
After the architecture parameter $\alpha$ is obtained, we can derive the final searched architecture following these steps like DARTS: 1) replace the mixed operation on each edge by   $o^{(i,j)}=\arg\max_{o\in\mathcal{O}, o\neq zero}p_o^{(i,j)}$, 2) retain two edges from different predecessor nodes with the largest $\max_{o\in\mathcal{O}, o\neq zero}p_o^{(i,j)}$.

\subsection{The Co-adaption Problem and Matthew Effect}
\begin{figure}[t]
	\vspace{-5pt}
	\centering
	\subfigure[CIFAR-10 first cell]{\includegraphics[width=.46\linewidth]{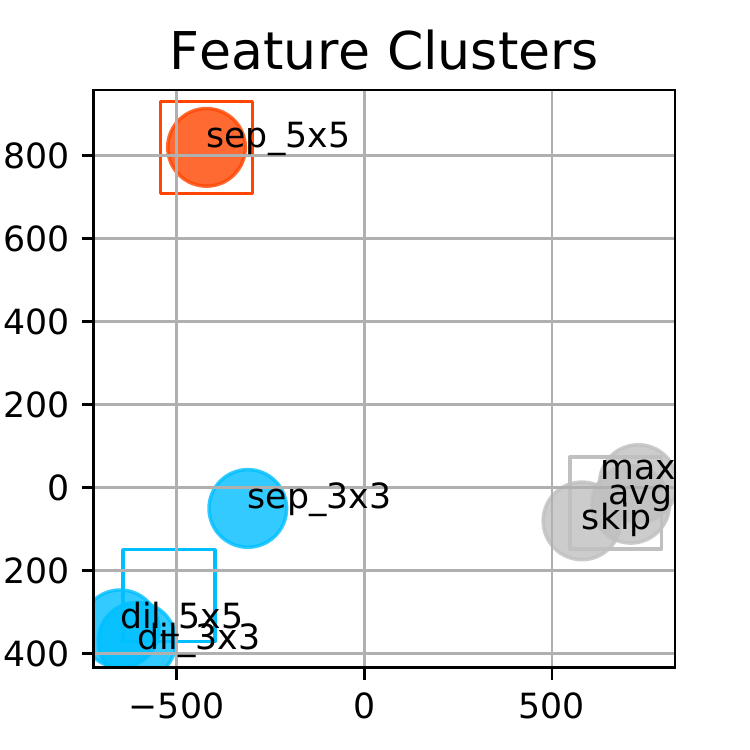}}
	%\hspace{5pt}
	\subfigure[CIFAR-10 last cell]{\includegraphics[width=.46\linewidth]{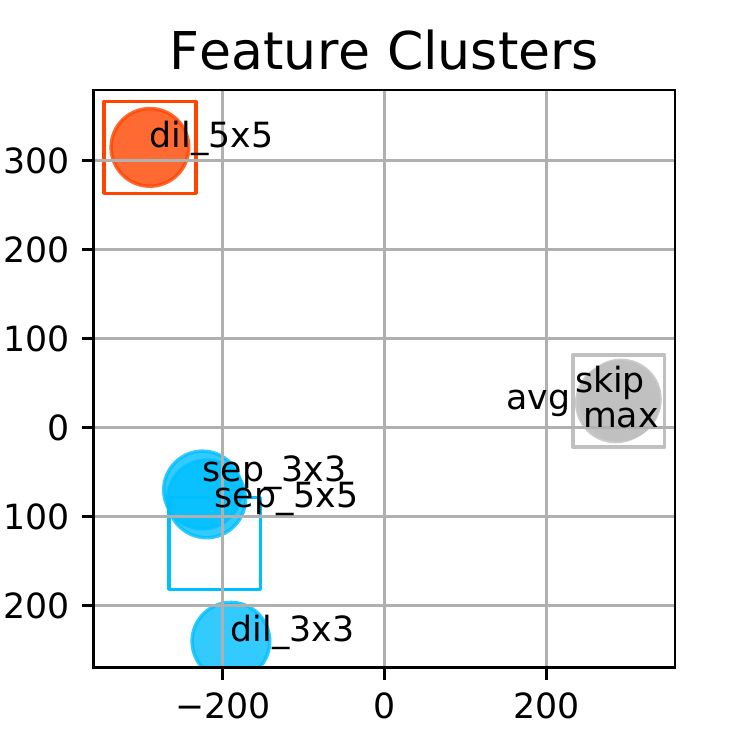}}
	%\hspace{-6pt}
	\subfigure[CIFAR-100 first cell]{\includegraphics[width=.46\linewidth]{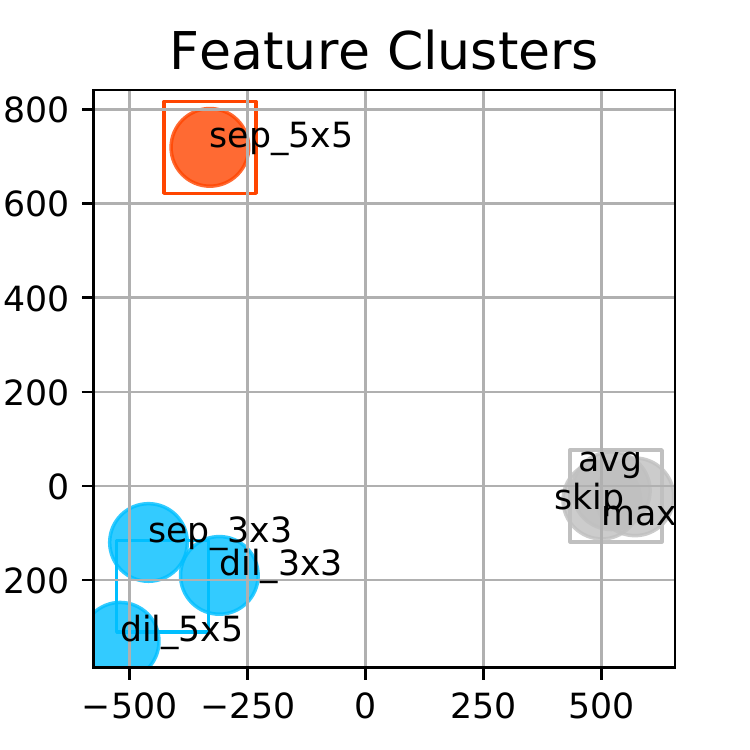}}
	%\hspace{-6pt}
	\subfigure[CIFAR-100 last cell]{\includegraphics[width=.46\linewidth]{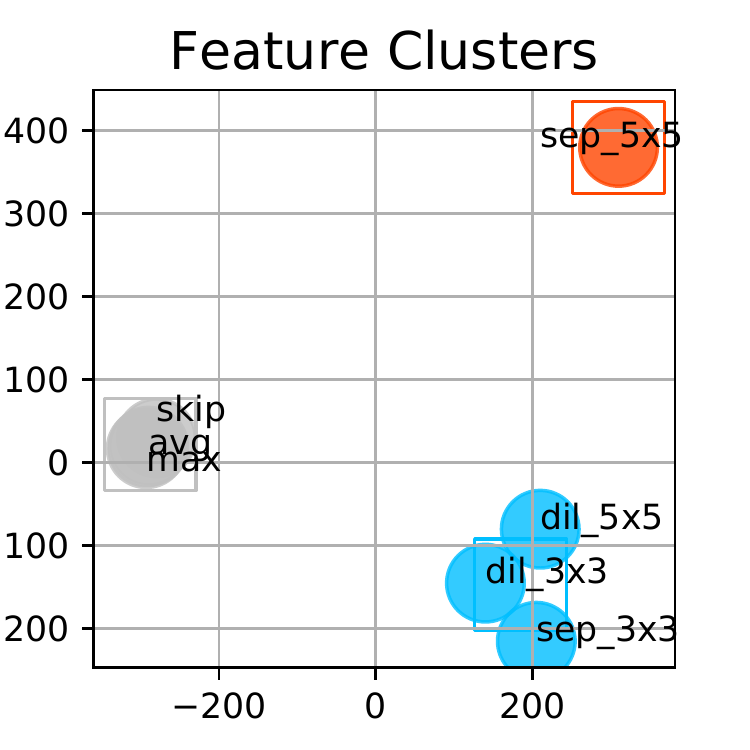}}
	%\vspace{-14pt}
	\caption{\small{Feature clusters generated from different operations on edge $E_{(0,2)}$: following \protect\cite{li2019stacnas}, we randomly select 1000 data, from CIFAR-10 and CIFAR-100 respectively, to generate the feature mappings, and apply K-means to cluster these mappings into 3 clusters to show the similarities between them, and finally use PCA to get a two-dimensional illustration. In (a) and (c) we select the edge in the first cell of the one-shot model, and in (b) and (d) we select the edge in the last cell. }}\label{fig:cluster}
	%\vspace{-50pt}
	% We use 1000 data randomly sampled from CIFAR-10 and CIFAR-100 to generate the feature mappings in the one-level DARTS. Under all the cases, we can observe a single green point that represented as one of the parameterized operation, shifts far away from other parameterized ones, and we can also see the non-parameterized points locate very close to each other, hard to distinguish.
\end{figure}

To  explore the co-adaption phenomenon, we visualize the clustering of  the feature maps generated from all the seven operations (except \textit{zero}) on edge $E_{(0,2)}$ in Figure \ref{fig:cluster}. We find that similar operations like convolutions with different kernel sizes generally produce similar feature maps, which means they serve as similar functions in the system. However, it is also common that convolutions with larger kernel size and more parameters stand far away from the other convolutions, which suggests that they are expressing something different from the others, while these standing out operations are always getting lower architecture value in the system. Then we check the stand-alone accuracy of these operations by training the corresponding architecture separately, and we find that large kernels usually perform better if trained properly, which contradicts the score they obtained from the DARTS system.  This suggests that the system suffers from the co-adaption problem that those with fewer parameters would be trained maturely earlier and express the desired function more quickly,  causing that \textit{5x5 conv}s rarely have chance to express the desired function. This also causes their $\alpha$ to be smaller and less gradient to backward to them and consequently causes the Matthew Effect.

In this work, we will introduce operation dropout to DARTS  based on one-level optimization to explore a more general search scheme, which actually unifies the existing differentiable methods and finds the best strategy for weight-sharing in the DARTS framework.

\subsection{Grouped Operation Dropout}\label{sec:god}
In this part,  we propose a simple yet effective search scheme called \textit{Grouped Operation Dropout} to break the correlation in the weight-sharing supernet.

Specifically, during the search stage, for  each edge, we randomly and independently select a subset of the operations, and zero out their outputs, to make their $\alpha$ and $w$ not  updated during back-propogation. Such a strategy mitigates the co-adaption among the parameterized operations since the under fitted operations have more chance to play an important role during the one-shot model training so that it can better fit the training target and benefit the $\alpha$'s learning. %The multiple collinearity problem in correlated operations is also alleviated because dropout is essentially a regularization technique that reduce the degree of freedom of $\alpha$ \cite{srivastava2014dropout}.

In practice, we partition the eight operations into two groups according to whether they are learnable, i.e. one parameterized group $\mathcal{O}_{p}$ containing all the convolutional operations, and one non-parameterized group $\mathcal{O}_{np}$ involving the remains. During the entire search procedure, for group $\mathcal{O}_{p}$ as an example, we fix the probability of each operation to be dropped as $p_d=r^{1/|\mathcal{O}_p|}$, where $0<r<1$ is a hyperparameter called drop path rate. Note that $|\mathcal{O}_{p}|=|\mathcal{O}_{np}|=4$, and the hyperparameter $r$ denotes the probability of disabling all the operations in $\mathcal{O}_p$ and $\mathcal{O}_p$ in the DARTS search space. For example, if we set $r=3\times10^{-5}$, then $p_d=r^{1/4}=0.074$. Additionally, we also enforce at least one operation to remain in each group to further stabilize the training, which is realized by resampling if the operations on some edge happen to be all dropped. During the backpropagation period, the $w$ and $\alpha$ of the dropped operations will receive no gradient. By enforcing to keep at least one operation in each group, the equivalent function of an edge is always a mixture of learnable and non-learnable operations, resulting in a relatively stable training environment for the architecture parameters $\alpha$.

%There are two major advantages of grouped dropout. The first is for each group we can employ different dropout rates during the search stage for achieving different objectives, like reducing the co-adaption problem in $\mathcal{O}_p$ and the multiple collinearity problem in $\mathcal{O}_{np}$. The second is by enforcing to keep at least one operation in each group, the equivalent function of each edge is always a mixture of learnable and non-learnable operations, resulting in a relatively stable training environment for the architecture parameters $\alpha$.

Note that operation dropout in one-level DARTS essentially unifies  most existing differentiable NAS appoaches: DARTS \cite{liu2018darts} updates all the operations on an edge at once, which corresponds to the $p_d=0$ case; SNAS \cite{xie2018snas} and ProxylessNAS \cite{cai2018proxylessnas} once samples only one and two operations to update, corresponding to  $p_d=0.875$ and $p_d=0.75$ respectively in expectation. We will later show that all of them are actually inferior to the best $p_d$ we find.

\paragraph{$\alpha$-Adjust: Prevent Passive Update.} Note that in DARTS, we measure the contribution $p_o^{(i,j)}$ of a certain operation $o^{(i,j)}$ on the edge $E_{(i,j)}$ via a softmax over all the learnable architecture parameters $\alpha_o^{(i,j)}$, as  in Equation \eqref{con:softmax}. As a result, the contribution $p_o^{(i,j)}$ of the dropped operations that do not receive any gradient during the backward pass, will get changed even though their corresponding $\alpha_{o}^{(i,j)}$ remain the same. In order to prevent the passive update of the dropped operations' $p_o^{(i,j)}$, we need to adjust the value of each $\alpha_o^{(i,j)}$ after applying the gradient. Our approach is to solve for an additional term $x$ according to:
\begin{align}
	\frac{\sum_{o\in\mathcal{O}_d}\exp(\alpha_o^{old})}{\sum_{o\in\mathcal{O}_k}\exp(\alpha_o^{old})} = \frac{\sum_{o\in\mathcal{O}_d}\exp(\alpha_o^{new})}{\sum_{o\in\mathcal{O}_k}\exp(\alpha_o^{new}+x)} \label{con:adjust}
\end{align}
where we omitted the subscript $(i,j)$,  $\mathcal{O}_d$ \& $\mathcal{O}_k$ refer to the operation sets that are dropped \& kept on edge $E_{(i,j)}$, $\alpha_o^{old}$ \& $\alpha_o^{new}$ means the value before \& after backpropagation. With the additional term $x$ to adjust the value of $\alpha_o^{new}$ for $o\in\mathcal{O}_k$, the contribution $p_o^{(i,j)}$ for $o\in\mathcal{O}_d$ remains unchanged. Note that $\alpha_o^{old}=\alpha_o^{new}$ for $o\in\mathcal{O}_d$, by solving Equation \eqref{con:adjust} we get:
\begin{align}
	x = \ln\left[\frac{\sum_{o\in\mathcal{O}_k}\exp(\alpha_o^{old})}{\sum_{o\in\mathcal{O}_k}\exp(\alpha_o^{new})}\right]. \label{con:x_equa}
\end{align}

\paragraph{Partial-Decay: Prevent Unnecessary Weight Decay.}

L2 regularization is employed during the search stage of original DARTS, and we also find it useful in one-level optimization. However, when applied with dropout, the parameters of $\mathcal{O}_d$ will be regularized even when they are dropped. So in our implementation, we apply the L2 weight decay only to those $w$ and $\alpha$ belonging to $\mathcal{O}_k$ to prevent the over-regularization.

\section{Related Work}
In their scientific investigation work, \cite{bender2018understanding} explored path level dropout during the training of the supernet for NAS, concluding that a proper drop path rate is desired for reducing the co-adaption problem and maintaining the stability of the training. However, their findings are largely neglected in the differentiable NAS community, where most of the current works including DARTS, P-DARTS and StacNAS train all operations simultaneously.  \cite{cai2018proxylessnas,guo2019single} train the supernet by sampling one path with probability proportional to architecture or uniform distribution respectively, which is equivalent to drop path rate as $\frac{N-1}{N}$, where N is the total number of operations. However, our empirical results show that such a high drop path rate is not the best choice of training the supernet of DARTS search space: the system is instable where convolution is sometimes followed by pooling and sometimes a convolution.

In this work, we introduce dropout to the DARTS framework. By adopting a properly tuned drop path rate and leveraging the operation grouping and one-level optimization proposed by \cite{li2019stacnas}, we show that we could further improve the most SOTA results achieved before.

\renewcommand\tabcolsep{3.5pt}

\begin{table}[t]
	\scriptsize
	\centering
	\begin{tabular}{cccccc}
		\hline
		\multicolumn{1}{c}{\multirow{2}{*}{\textbf{Architecture}}}  & \multicolumn{1}{c}{\multirow{2}{*}{\textbf{Test Error (\%)}}} & \textbf{Param} & \textbf{Search Cost} \\
		
		\multicolumn{1}{c}{}  & \multicolumn{1}{c}{} & \textbf{(M)} & \textbf{(GPU Days)} \\
		\hline
		NASNet-A~\cite{zoph2016neural} & $2.65$ & $3.3$ & $1800$  \\
		AmoebaNet-B~\cite{real2019regularized} & $2.55 \pm 0.05$ & $2.8$ & $3150$ \\
	%	PNAS~\cite{liu2018progressive}$^1$ & $3.41 \pm 0.09$ & $3.2$ & $225$ & SMBO \\
		ENAS~\cite{pham2018efficient} & $2.89$ & $4.6$ & $0.5$ \\
		\hline
		DARTS~\cite{liu2018darts} & $3.00$ & $3.3$ & $1.5$  \\
		SNAS~\cite{xie2018snas} & $2.85$ & $2.8$ & $1.5$  \\
		ProxylessNAS~\cite{cai2018proxylessnas}$^1$ & $2.08$& $5.7$ & $4$  \\
		P-DARTS~\cite{chen2019progressive} & $2.50$ & $3.4$ & $0.3$  \\
		DARTS+~\cite{liang2019darts+}$^2$ & $2.20(2.37 \pm 0.13)$ & $4.3$ & $0.6$  \\
		StacNAS~\cite{li2019stacnas} & $2.33(2.48 \pm 0.08)$ & $3.9$ & $0.8$  \\
		ASAP~\cite{noy2019asap} & $2.49 \pm 0.04$ & $2.5$ & $0.2$  \\
		PC-DARTS~\cite{xu2019pc} & $2.57 \pm 0.07$ & $3.6$ & $0.1$  \\
		\hline
		\textbf{DropNAS}$^3$ & $\mathbf{2.26} (2.58 \pm 0.14)$ & $4.1$ & $0.6$  \\
		\textbf{DropNAS (Augmented)}{$^4$} & $\mathbf{1.88}$ & $4.1$ & $0.6$  \\
		\hline
	\end{tabular}
	\caption{Performance of different architectures on CIFAR-10.  ProxylessNAS$^1$ uses a search space different from DARTS. DARTS+$^2$ trains the evaluation model for 2,000 epochs, while others just train 600 epochs. Our DropNAS$^3$ reports both the mean and standard deviation with eight seeds, \textbf{by keeping training epochs or channels the same with the original DARTS for a fair comparison}. DropNAS (Augmented)$^4$ denotes training with AutoAugment and 1,200 epochs.}
	\label{tab:cifar10_results}
\end{table}

\begin{table}[t]
	\scriptsize
	\centering
	\begin{tabular}{cccccc}
		\hline
		\multicolumn{1}{c}{\multirow{2}{*}{\textbf{Architecture}}}  & \multicolumn{1}{c}{\multirow{2}{*}{\textbf{Test Error (\%)}}} & \textbf{Param} & \textbf{Search Cost}  \\
		
		\multicolumn{1}{c}{}  & \multicolumn{1}{c}{} & \textbf{(M)} & \textbf{(GPU Days)} \\
		\hline
		DARTS~\cite{liu2018darts}$^1$ & $17.76$ & $3.3$ & $1.5$  \\
		P-DARTS~\cite{chen2019progressive} & $15.92$ & $3.6$ & $0.3$  \\
		DARTS+~\cite{liang2019darts+}$^2$ & $14.87(15.45 \pm 0.30)$ & $3.9$ & $0.5$  \\
		StacNAS~\cite{li2019stacnas} & $15.90(16.11 \pm 0.2)$ & $4.3$ & $0.8$  \\
		ASAP~\cite{noy2019asap}$^1$ & $15.6$ & $2.5$ & $0.2$  \\
		\hline
		\textbf{DropNAS}$^3$ & $16.39 (16.95 \pm 0.41)$ & $4.4$ & $0.7$  \\
		\textbf{DropNAS (Augmented)}{$^4$} & $\mathbf{14.10}$ & $4.4$ & $0.7$  \\
		\hline
	\end{tabular}
	\caption{Results of different architectures on CIFAR-100. The results  denoted with $^1$ use the architectures found on CIFAR-10. The subscript $^2,^3$ and $^4$ have the same meaning as in Table \ref{tab:cifar10_results}.}
	\label{tab:cifar100_results}
	%\vspace{-10pt}
\end{table}

\section{Benchmark}

\subsection{Datasets}

To benchmark our grouped operation dropout algorithm, extensive experiments are carried out on CIFAR-10, CIFAR-100 and ImageNet. 

Both the CIFAR-10 and CIFAR-100 datasets contain 50K training images and 10K testing images, and the resolution of each image is $32\times 32$. All the images are equally partitioned into 10/100 categories in CIFAR-10/100.

ImageNet is a much larger dataset consisting of 1.3M images for training and 50K images for testing, equally distributed among 1,000 classes. In this paper, we use ImageNet to evaluate the transferability of our architectures found on CIFAR-10/100. We follow the conventions of \cite{liu2018darts} that consider the mobile setting to fix the size of the input image to $224\times224$ and limit the multiply-add operations to be no more than 600M.

\subsection{Implementation Details}

\paragraph{Architecture Search} \label{sec:search}

As we have mentioned before, we leverage the DARTS search space with the same eight candidate operations. Since we use one-level optimization, the training images do not need to be split for another validation set, so the architecture search is conducted on CIFAR-10/100 with all the training images on a single Nvidia Tesla V100. We use 14 cells stacked with 16 channels to form the one-shot model, train the supernet for 76 epochs with batch size 96, and pick the architecture discovered in the final epoch. The model weights $w$ are optimized by SGD with initial learning rate 0.0375, momentum 0.9, and weight decay 0.0003, and we clip the gradient norm of $w$ to be less than 3 for each batch. The architecture parameters $\alpha$ are optimized by Adam, with initial learning rate 0.0003, momentum (0.5, 0.999) and weight decay 0.001. Drop path rate $r$ is fixed to $3\times10^{-5}$.

\begin{figure}[t]
	\vspace{-10pt}
	\centering
	\subfigure{\includegraphics[width=.49\linewidth]{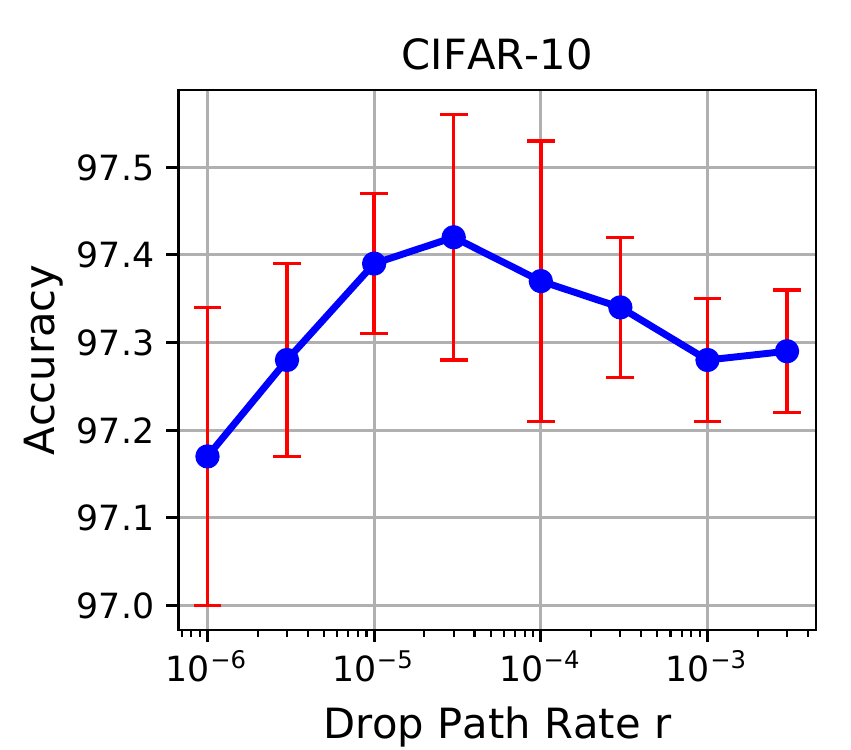}}
	%\hspace{5pt}
	\subfigure{\includegraphics[width=.49\linewidth]{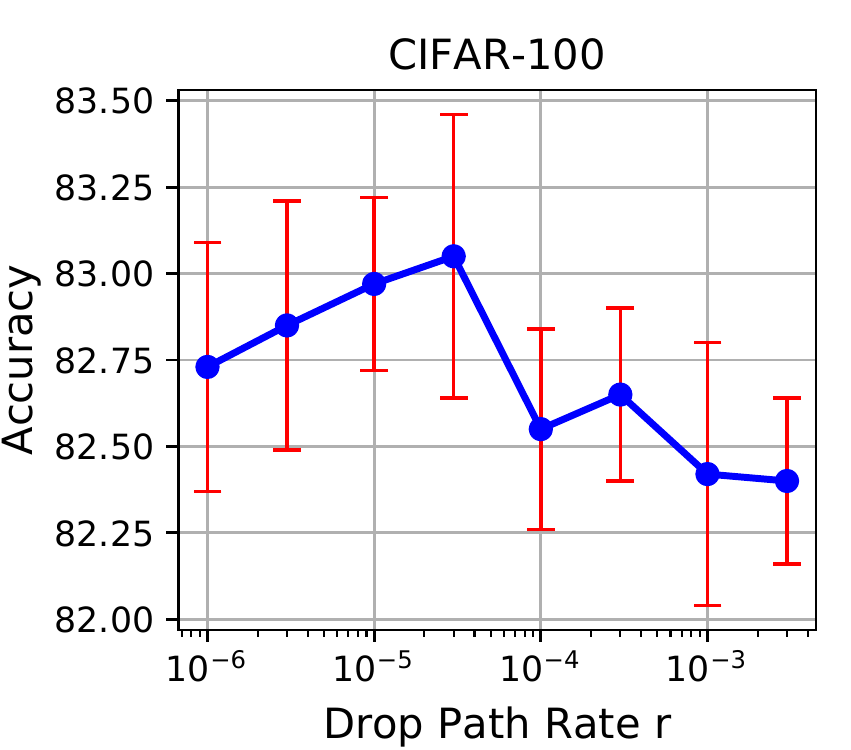}}
	%\hspace{-6pt}
	%\hspace{-14pt}
	\caption{\small{The impact of drop path rate reflected on the stand-alone model accuracy. The red error bars show the standard deviation of 8 repeated experiments. }}\label{fig:rate}
	% The best rate $r=3\times10^{-5}$ is transferable on different datasets.
	\vspace{-10pt}
\end{figure}

\begin{figure}[t]
	\vspace{-3pt}
	\centering
	\subfigure[CIFAR-10 normal cell]{\includegraphics[width=.49\linewidth]{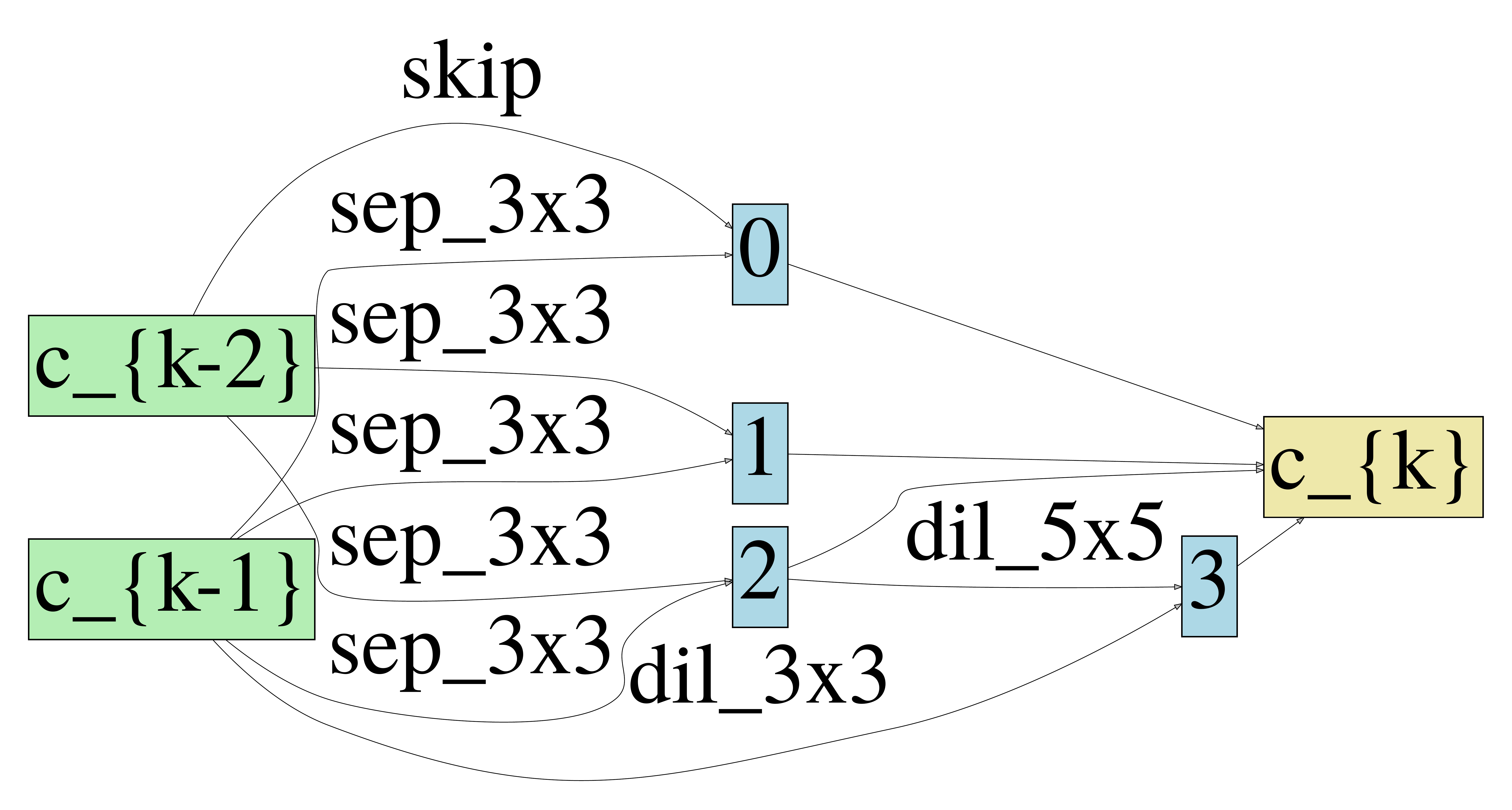}}
	%\vspace{-10pt}
	\subfigure[CIFAR-100 normal cell]{\includegraphics[width=.50\linewidth]{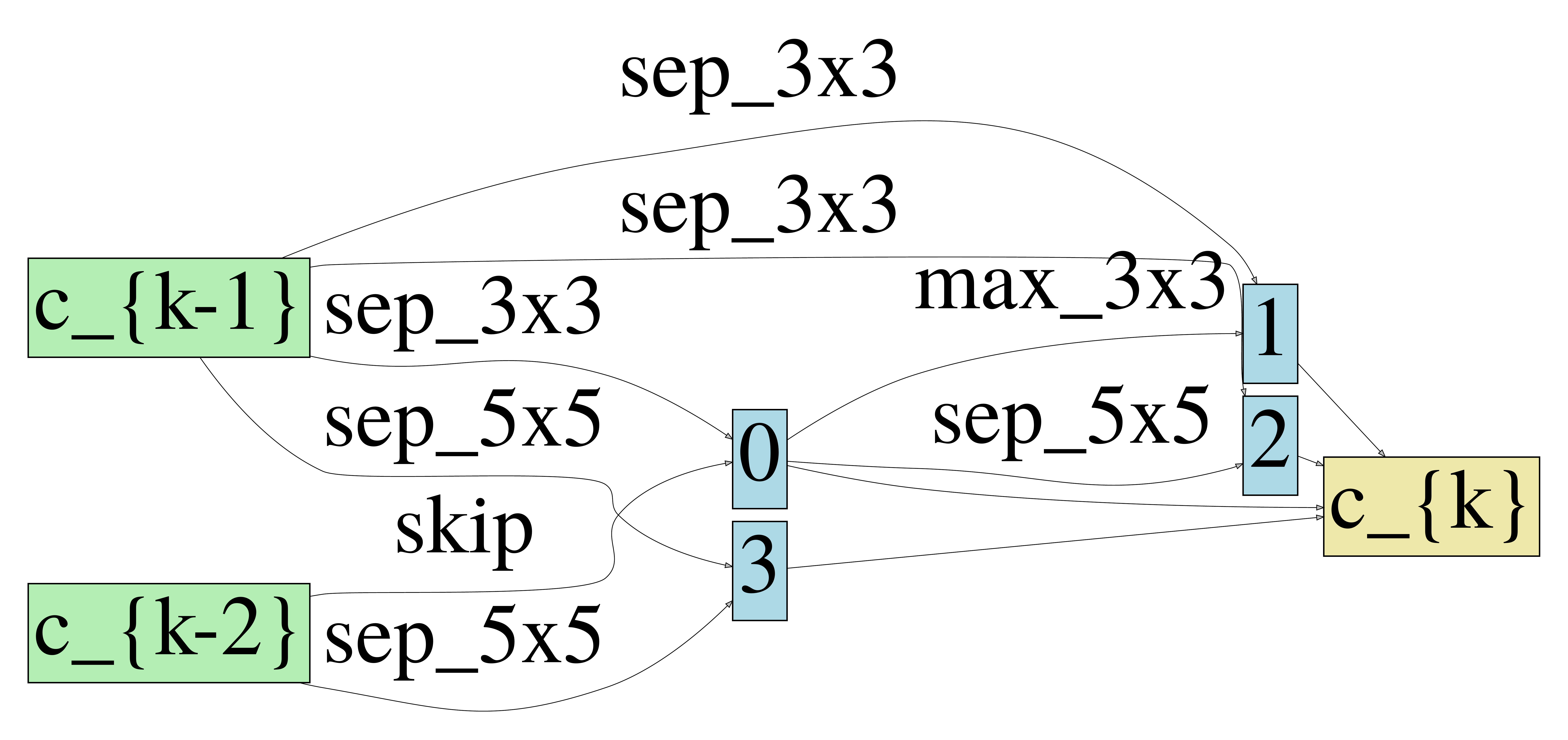}}
	%\vspace{-10pt}
	\subfigure[CIFAR-10 reduction cell]{\includegraphics[width=.99\linewidth]{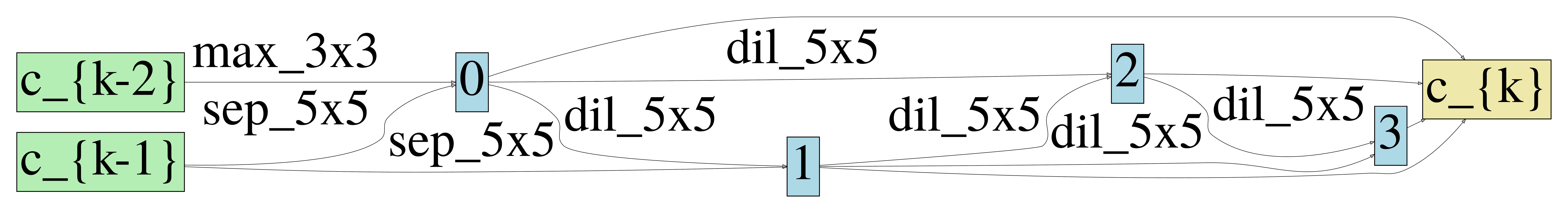}}
	%\vspace{-6pt}
	\subfigure[CIFAR-100 reduction cell]{\includegraphics[width=.99\linewidth]{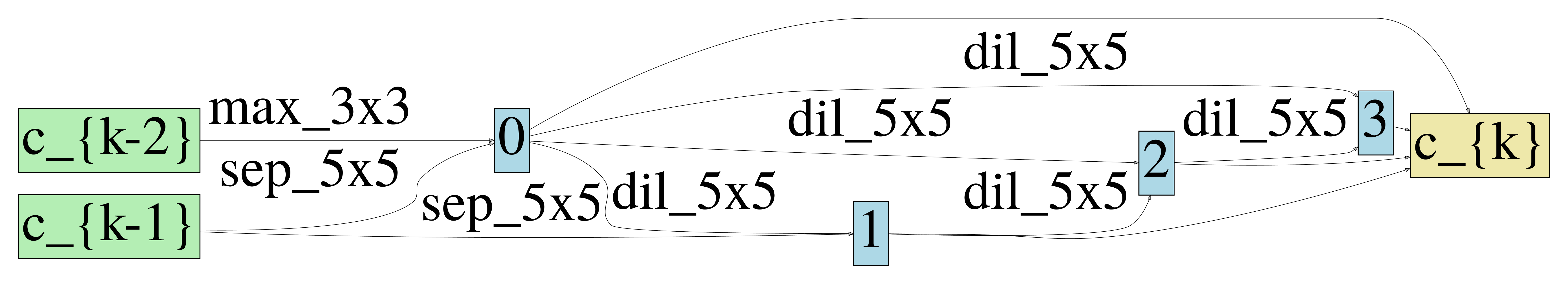}}
	%\hspace{-14pt}
	\caption{\small{The found architectures on CIFAR-10 and CIFAR-100}}\label{fig:arch}
	%\vspace{-10pt}
\end{figure}

\paragraph{Architecture Evaluation}

On CIFAR-10 and CIFAR-100, to fairly evaluate the discovered architectures, neither the initial channels nor the training epochs are increased for the evaluation network, compared with DARTS. 20 cells are stacked to form the evaluation network with 36 initial channels. The network is trained on a single Nvidia Tesla V100 for 600 epochs with batch size 192. The network parameters are optimized by SGD with learning rate 0.05, momentum 0.9 and weight decay 0.0003, and the gradient is clipped in the same way as in the search stage. The data augmentation method Cutout and an auxiliary tower with weight 0.4 are also employed as in DARTS. To exploit the potentials of the architectures, we additionally use AutoAugment to train the model for 1,200 epochs. The best architecture discovered are represented in \fig{fig:arch} and their evaluation results are shown in Table \ref{tab:cifar10_results} and \ref{tab:cifar100_results}. We can see that the best architecture discovered by DropNAS achieves the SOTA test error $2.26\%$ on CIFAR-10, and on CIFAR-100 DropNAS still works well compared to DARTS, and largely surpasses the one-level version which prefers to ending up with many \textit{skip-connect} in the final architecture if directly searched on CIFAR-100.

To test the transferability of our selected architectures, we adopt the best architectures found on CIFAR-10 and CIFAR-100 to form a 14-cell, 48-channel evaluation network to train on ImageNet. The network is trained for 600 epochs with batch size 2048 on 8 Nvidia Tesla V100 GPUs, optimized by SGD with initial learning rate 0.8, momentum 0.9, weight decay $3\times 10^{-5}$, and gradient clipping 3.0. The additional enhancement approaches that we use include AutoAugment, mixup, SE module, auxiliary tower with loss weight 0.4, and label smoothing with $\epsilon=0.1$. Table \ref{tab:imagenet_results} shows that the architecture found by DropNAS is transferable and obtains encouraging result on ImageNet.

\begin{table}[t]
    %\vspace{-3pt}
	\centering
	\scriptsize
	\begin{tabular}{cccccc}
		\hline
		\multirow{2}{*}{\textbf{Architecture}} & \multicolumn{2}{c}{\textbf{Test Err. (\%)}} & \textbf{Params}  & \textbf{Search} & \\
		\cline{2-3}
		& \textbf{Top-1} & \textbf{Top-5} & \textbf{(M)} & \textbf{Days} &   \\
		\hline
		%ShuffleNet-V2 (2x)~\cite{ma2018shufflenet} & $25.1$ & - & $7.4$ & - & manual \\
		%\hline
		NASNet-A~\cite{zoph2016neural} & $26.0$ & $8.4$ & $5.3$ & $1800$  \\
		%AmoebaNet-C~\cite{real2019regularized} & $24.3$ & $7.6$ & $6.4$ & $3150$ & RL \\
		%PNAS~\cite{liu2018progressive} & $25.8$ & $8.1$ & $5.1$ & $588$ & SMBO \\
		%MnasNet-92~\cite{tan2019mnasnet} & $25.2$ & $8.0$ & $4.4$ & - & RL \\
		EfficientNet-B0~\cite{tan2019efficientnet} & $23.7$ & $6.8$ & $5.3$ & -  \\
		\hline
		DARTS~\cite{liu2018darts} & $26.7$ & $8.7$ & $4.7$ & $4.0$  \\
		SNAS (mild)~\cite{xie2018snas} & $27.3$ & $9.2$ & $4.3$ & $1.5$  \\
		ProxylessNAS~\cite{cai2018proxylessnas}$^\dag$$^*$ & $24.9$ & $7.5$ & $7.1$ & $8.3$  \\
		P-DARTS (C10)~\cite{chen2019progressive} & $24.4$ & $7.4$ & $4.9$ & $0.3$  \\
		ASAP~\cite{noy2019asap} & $26.7$ & - & - & $0.2$  \\
		XNAS~\cite{nayman2019xnas} & $24.0$ & - & $5.2$ & $0.3$  \\
		PC-DARTS~\cite{xu2019pc}$^\dag$ & $24.2$ & $7.3$ & $5.3$ & $3.8$  \\
		ScarletNAS\cite{chu2019scarletnas}$^{\dag}$$^*$ & $23.1$ & $6.6$ & $6.7$ & $10$ \\
		DARTS+\cite{liang2019darts+}$^{\dag}$ & $23.9$ & $7.4$ & $5.1$ & $6.8$  \\
		StacNAS\cite{li2019stacnas}$^{\dag}$ & $24.3$ & $6.4$ & $5.7$ & $20$  \\
		Single-Path NAS \cite{stamoulis2019single}$^{\dag}$$^*$ & $25.0$ & $7.8$ & $-$ & $0.16$  \\
		\hline
		\textbf{DropNAS (CIFAR-10)} & $\mathbf{23.4}$ & $\mathbf{6.7}$ & $5.7$ & $ 0.6 $  \\
		\textbf{DropNAS (CIFAR-100)} & $\mathbf{23.5}$ & $\mathbf{6.8}$ & $6.1$ & $ 0.7 $  \\
		\hline
	\end{tabular}
	\caption{Results of different architectures on ImageNet. The results denoted with $^\dag$ use the architectures directly searched on ImageNet, and those denoted with $^*$ use the backbone different from DARTS.}
	\label{tab:imagenet_results}
\end{table}

\begin{figure}[t]
	\vspace{-6pt}
	\centering
	\subfigure[CIFAR-10 first]{\includegraphics[width=.46\linewidth]{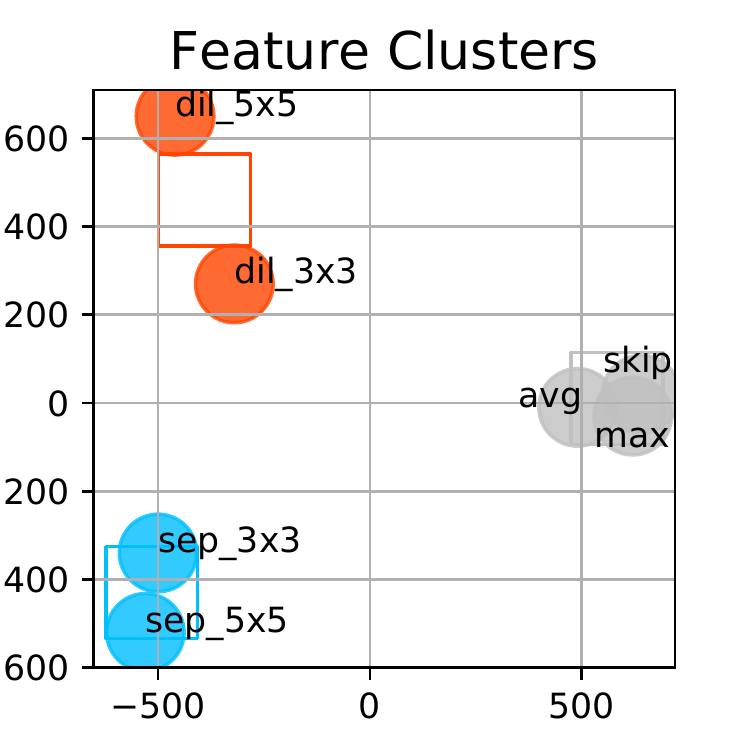}}
	%\hspace{5pt}
	\subfigure[CIFAR-10 last]{\includegraphics[width=.46\linewidth]{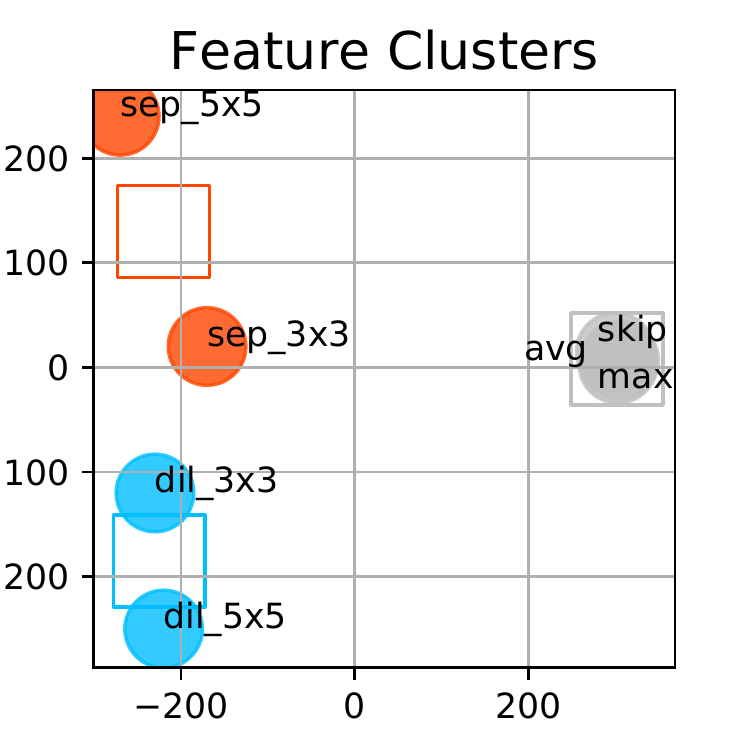}}
	%\hspace{-6pt}
	\subfigure[CIFAR-100 first]{\includegraphics[width=.46\linewidth]{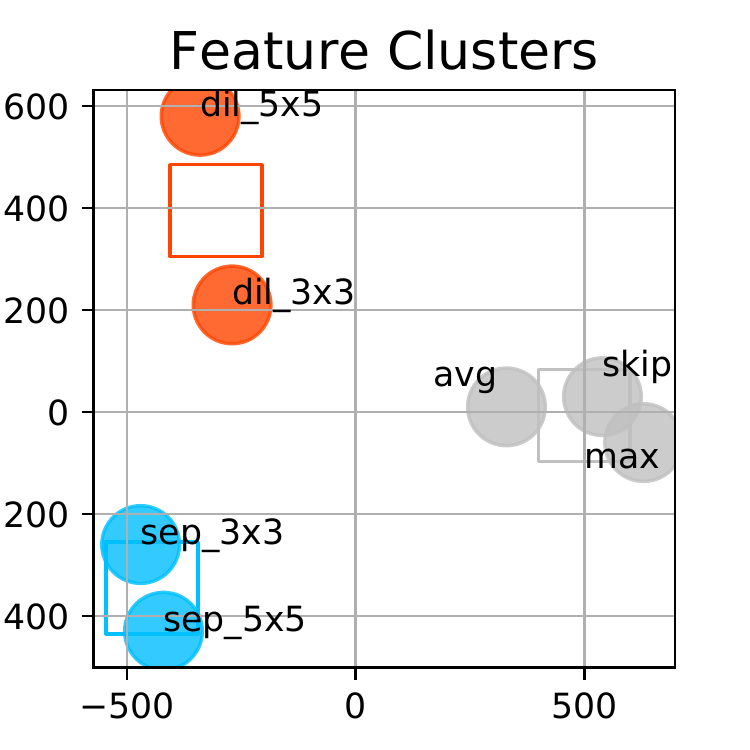}}
	%\hspace{-6pt}
	\subfigure[CIFAR-100 last]{\includegraphics[width=.46\linewidth]{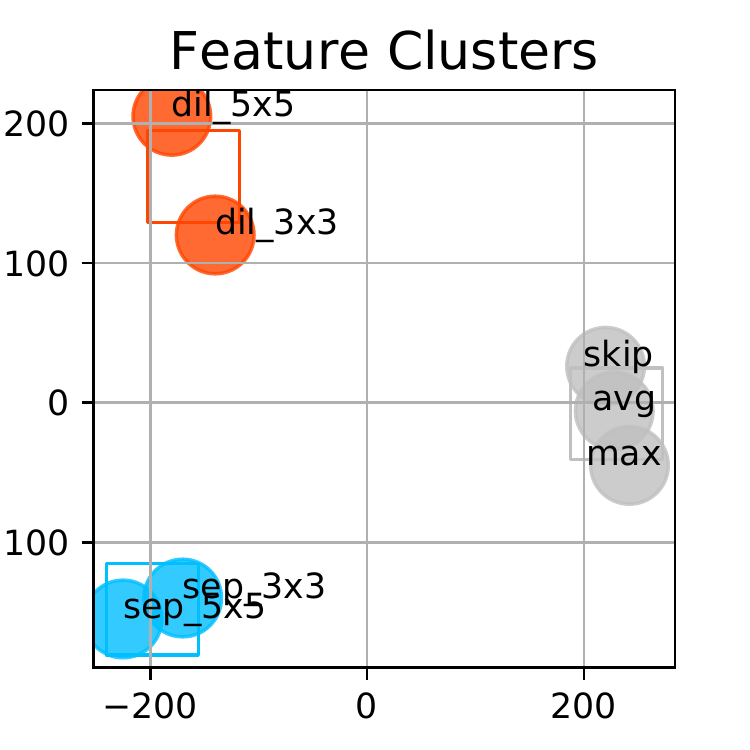}}
	%\hspace{-14pt}
	\caption{\small{Feature clusters of DropNAS on $E_{(0,2)}$}} \label{fig:cluster_drop}
	\vspace{-10pt}
\end{figure}

\section{Diagnostic Experiments}

\subsection{Impact of Drop Path Rate}

In DropNAS we introduce a new hyperparameter, i.e. the drop path rate $r$, whose value has a strong impact on the results since a higher drop path rate results in a lower correlation between the operations. To demonstrate its significance, we repeat the search and evaluation stages with varying drop path rates and report the stand-alone model accuracy in \fig{fig:rate}. The best results are achieved when $r=3\times10^{-5}$ on both datasets, which indicates that the found best drop path rate is transferable to different datasets. Note that $p_d$ is just $0.074$ when $r=3\times10^{-5}$, so the other cases like $p_d=0.875, 0.75$ or $0$ are all inferior to it, which correspond to the search scheme of SNAS, ProxylessNAS and DARTS, respectively.

\subsection{Feature Clusters in DropNAS}

For comparison we again draw the feature clusters in DropNAS with $r=3\times10^{-5}$, following the same way in \fig{fig:cluster}. The results are plotted in \fig{fig:cluster_drop}.

It is significant that the point of parameterized operation no longer shifts away from its similar partner, and there is no cluster containing only one single point anymore. So we claim that the severe co-adaption problem existing in the one-level DARTS has been greatly reduced by DropNAS.

%Multiple collinearity problems can be naturally alleviated by applying DropNAS to reduce the degree of freedom in $\alpha$'s learning. To our surprise, the points of non-parameterized operations also become more scattered compared to \fig{fig:cluster}, which may due to more distinguishable feature mappings are learned in DropNAS.

\subsection{Performance on Other Search Space}

We are also interested in the adaptability of DropNAS in other search spaces. We purposely design two search spaces: in the first space we replace the original \textit{3x3 avg-pooling} and \textit{3x3 max-pooling} operations by \textit{skip-connect}; And in the second space we remove the \textit{3x3 avg-pooling} and \textit{3x3 max-pooling} operations in $\mathcal{O}_{np}$. We again search on CIFAR-10 and evaluate the found architectures, report the mean accuracy and standard deviations of eight repeated runs.

The results shown in Table \ref{tab:diff_space} demonstrates that DropNAS is robust across variants of the DARTS search spaces in different datasets.

%In the 3-skip space, the architectures discovered by one-level DARTS are full of convolutional operations without any \textit{skip-connect}, since the three same operations are highly correlated and disperse the significance of each other. DropNAS reduces such correlation and all the discovered architectures have at least one \textit{skip-connect}. And in the 1-skip search space, DropNAS again maintains the high performance. The number of \textit{skip-connect} found by one-level DARTS varies from 3 to 5 due to the correlation between similar convolutions.

\subsection{Impact of Drop Path Rates in Different Groups}

As we mentioned in Section \ref{sec:god}, one advantage of grouping in DropNAS is that we can apply different drop path rates to different operation groups. However, our architecture search is actually conducted with $r$ fixed to $3\times 10^{-5}$ for both $\mathcal{O}_p$ and $\mathcal{O}_{np}$. In fact, we have assigned $\mathcal{O}_p$ and $\mathcal{O}_{np}$ with different drop path rates around $3\times 10^{-5}$, and the results are shown in Table \ref{tab:diff_rates}, which means the best performance is achieved when the two groups share exactly the same rate.

\renewcommand{\arraystretch}{1.2}

\begin{table}[]
	\vspace{-5pt}
	\scriptsize
	\begin{tabular}{cccc}
		\hline
		\multicolumn{1}{c}{\multirow{2}{*}{\textbf{Dataset}}} & \multicolumn{1}{c}{\multirow{1}{*}{\textbf{Search}}} & \multicolumn{2}{c}{\textbf{Test Error (\%)}} \\
		\cline{3-4}
		\multicolumn{1}{c}{} & \multicolumn{1}{c}{\textbf{Space}} & \textbf{DropNAS} & \textbf{one-level DARTS} \\
		\hline
		\multicolumn{1}{c}{\multirow{3}{*}{CIFAR-10}} &  3-skip & 2.68$\pm$0.10   & 3.19$\pm$0.18              \\
		\multicolumn{1}{c}{} & 1-skip        & 2.67$\pm$0.11    & 2.85$\pm$0.12         \\
		\multicolumn{1}{c}{} & original                   & 2.58$\pm$0.14    & 2.90$\pm$0.16           \\
		\hline
		\multicolumn{1}{c}{\multirow{3}{*}{CIFAR-100}} & 3-skip & 16.97$\pm$0.35      & 18.00$\pm$0.34         \\
		\multicolumn{1}{c}{} & 1-skip                   & 16.47$\pm$0.19    & 17.73$\pm$0.25           \\
		\multicolumn{1}{c}{} & original                   & 16.95$\pm$0.41    & 17.27$\pm$0.36           \\
		\hline
	\end{tabular}
	\caption{\small{The performance of DropNAS and one-level DARTS across different search spaces on CIFAR-10/100.}}\label{tab:diff_space}
\end{table}

\begin{table}[]
    \vspace{-5pt}
	\scriptsize
	\centering
	\begin{tabular}{|l|l|l|l|}
		\hline
		\diagbox{$\mathcal{O}_{np}$}{$\mathcal{O}_{p}$} & $1\times10^{-5}$ &
		$3\times10^{-5}$        & $1\times10^{-4}$ \\
		\hline
		$1\times10^{-5}$ & 2.60$\pm$0.16   & 2.72$\pm$0.04          & 2.64$\pm$0.12   \\
		\hline
		$3\times10^{-5}$ & 2.64$\pm$0.11   & \textbf{2.58}$\pm$\textbf{0.14} & 2.69$\pm$0.05   \\
		\hline
		$1\times10^{-4}$ & 2.65$\pm$0.07   & 2.69$\pm$0.10          & 2.63$\pm$0.16   \\
		\hline
	\end{tabular}
	\caption{\small{The test error of DropNAS on CIFAR-10 when the operation groups $\mathcal{O}_{p}$ and $\mathcal{O}_{np}$ are applied with different drop path rates. The above results are obtained over 8 different seeds.}} \label{tab:diff_rates}
	
\end{table}

\subsection{Performance Correlation between Stand-Alone Model and Architecture Parameters}

\begin{figure}[t]
    %\vspace{-20pt}
	\centering
	\subfigure[on edge $E_{(0,2)}: 0.902$]{\includegraphics[width=.68\linewidth]{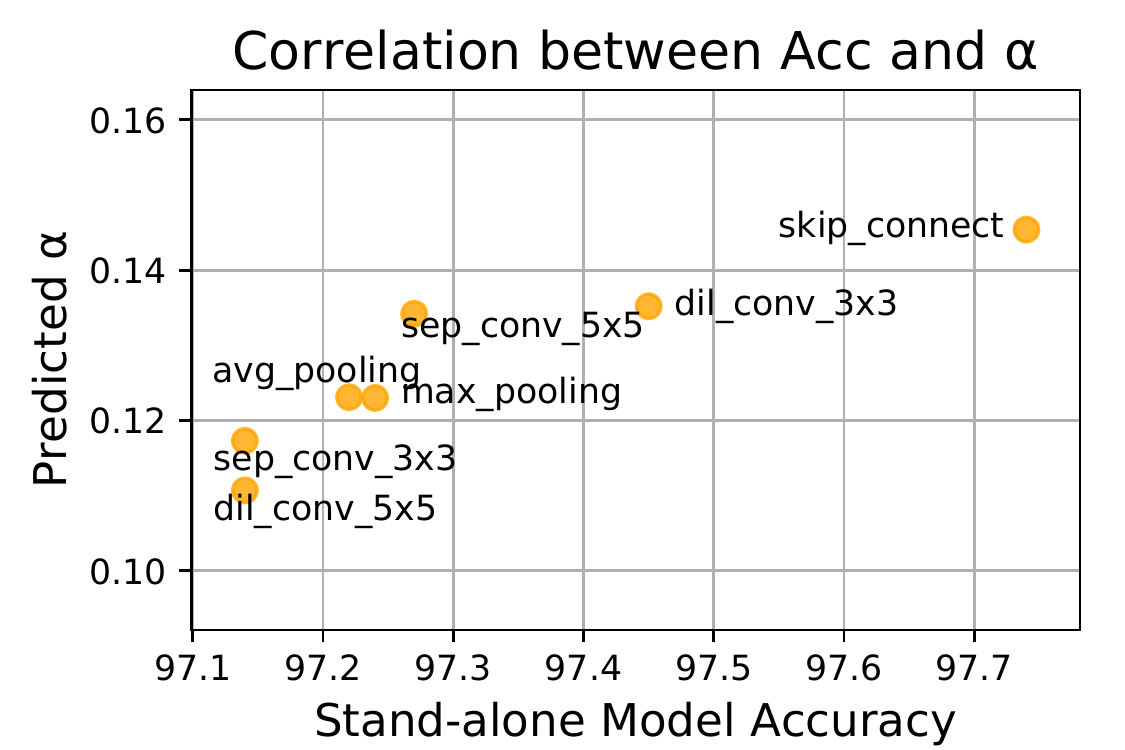}}
	%\hspace{5pt}
	\subfigure[on edge $E_{(4,5)}: 0.352$]{\includegraphics[width=.68\linewidth]{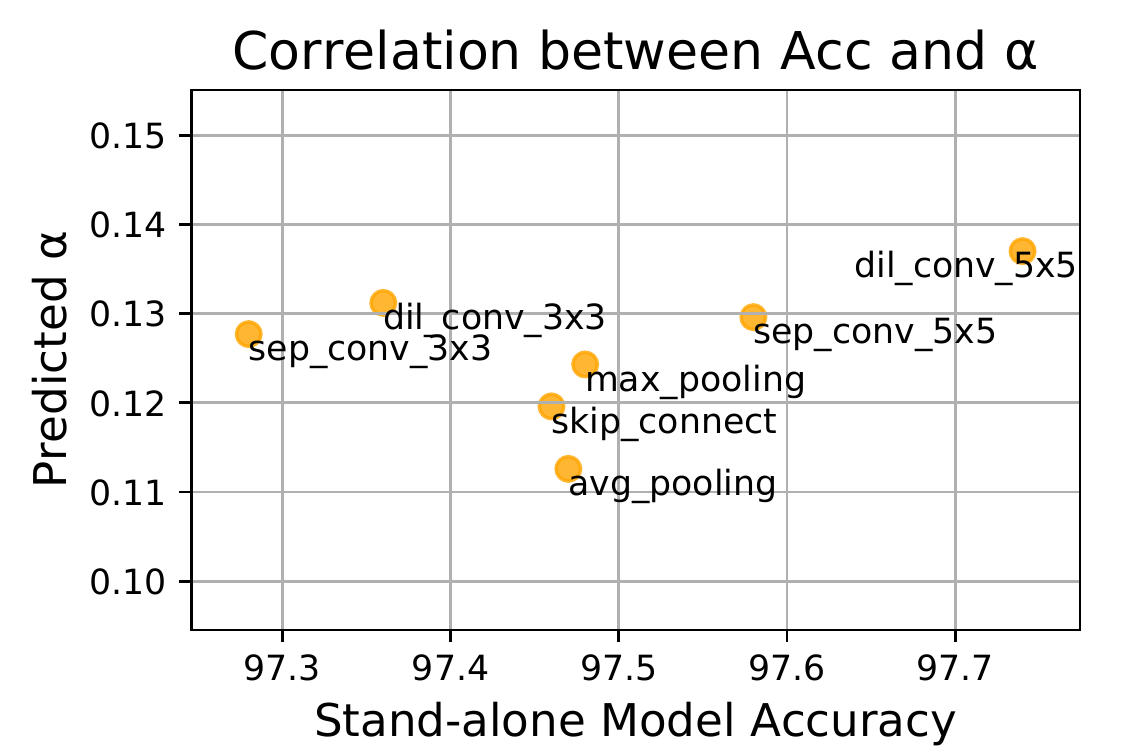}}
	\vspace{-5pt}
	\caption{\small{Correlation coefficients between the accuracy of stand-alone model and their corresponding $\alpha$. The results are obtained by first searching on CIFAR-10, figuring out the best architecture, then generating other 6 architectures by replacing the operation on edges $E_{(0,2)}$ and $E_{(4,5)}$ in the normal cell with other $o\in\mathcal{O}$, and finally the corresponding stand-alone models are trained from scratch.}}\label{fig:corr_acc}
	%\vspace{-10pt}
\end{figure}

DropNAS is supposed to break the correlation between the operations, so that the architecture parameters $\alpha$ can represent the real importance of each operation, and then we can easily select the best the architecture by ranking $\alpha$. \fig{fig:corr_acc} shows the correlation between the architectures and their corresponding $\alpha$ on two representative edges in normal cell, $E_{(0,2)}$ and $E_{(4,5)}$, which are the first and the last edge within the cell. We claim that the $\alpha$ learned by DropNAS has a vigorous representative power of the accuracy of the stand-alone model, since the correlation coefficient between them is 0.902 on $E_{(0,2)}$, largely surpassing that of DARTS (0.2, reported in \cite{li2019stacnas}), and 0.352 on $E_{(4,5)}$, where the choice of a specific operation is less significant.

\begin{table}[]
	\vspace{-5pt}
	\scriptsize
	\centering
	\begin{tabular}{ccccc}
		\multicolumn{1}{c}{\multirow{2}{*}{}} & \multicolumn{4}{c}{\textbf{Test Err. (\%)}} \\
		\cline{2-5}
		\multicolumn{1}{c}{} & \textbf{DropNAS} & \textbf{No $\alpha$-adjust} & \textbf{No partial-decay} & \textbf{No grouping} \\
		\hline
		\textbf{CIFAR-10} & \textbf{2.58}$\pm$\textbf{0.14} & 2.75$\pm$0.08 & 2.71$\pm$0.06 & 2.74$\pm$0.11
		\\
		\textbf{CIFAR-100} & \textbf{16.95}$\pm$\textbf{0.41} & 17.40$\pm$0.22 & 17.62$\pm$0.37 & 17.98$\pm$0.33
		\\
		\hline
	\end{tabular}
	\caption{\small{Ablation study on CIFAR-10/100, averaged over 8 runs.}}\label{tab:ablation}
	\vspace{-5pt}
\end{table}

\subsection{Ablation Study}

To show the techniques we proposed in Section \ref{sec:god} really improve the DropNAS performance, we further conduct experiments for DropNAS with each of the techniques disabled. The results in Table \ref{tab:ablation} show that each component of DropNAS is indispensable for achieving good performance.

\section{Conclusion}

We propose DropNAS, a grouped operation dropout method for one-level DARTS, that greatly improves the DARTS performance over various benchmark datasets. We explore the co-adaptation problem of DARTS and present empirical evidence about how DARTS performance is degraded by this problem. It should be noticed that various differentiable NAS approaches are unified in our DropNAS framework, but fail to match the best drop path rate we find. Moreover, the found best drop path rate of DropNAS is transferable in different datasets and variants of DARTS search spaces, demonstrating its strong applicability in a wider range of tasks.

%% The file named.bst is a bibliography style file for BibTeX 0.99c
\bibliographystyle{named}
\bibliography{ijcai20}

\end{document}